\documentclass{article}
\usepackage{spconf}
\usepackage{adjustbox, amsmath, amssymb, graphicx, multirow, booktabs}
\usepackage{algorithm}
\usepackage{algpseudocode}
\usepackage{url}
\usepackage{pdfpages}
\usepackage{subfigure}
\usepackage{IEEEtrantools}
\makeatletter
\providecommand\add@text{}
\newcommand\tagaddtext[1]{%
  \gdef\add@text{#1\gdef\add@text{}}}%
\renewcommand\tagform@[1]{%
  \maketag@@@{\llap{\add@text\quad}(\ignorespaces#1\unskip\@@italiccorr)}%
}
\makeatother

\usepackage{pgfplots}
\pgfplotsset{grid style={dashed,gray}}
\pgfplotsset{compat=1.12}
\usepackage{tikz}
\usetikzlibrary{shapes.geometric}
\usetikzlibrary{shapes.misc,shadows}
\usetikzlibrary{positioning} 
\usetikzlibrary{arrows} 
\usetikzlibrary{arrows.meta}
\tikzset{%
    >={Latex[width=1mm,length=1mm]},
    base/.style = {
        rectangle, rounded corners, draw=black,
        minimum width=2cm, minimum height=.4cm,
        text centered, font=\tiny},
    acoustic_model/.style = {base, fill=red!15},
    language_model/.style = {base, fill=cyan!20},
    joint/.style = {base, fill=yellow!15},
    io/.style = {base, fill=none, draw=none, minimum width=0cm},
    data/.style = {
        rectangle, draw, align=center, left color=blue!20, right color=white,
        minimum width=0.5cm, minimum height=0.5cm},
    block/.style ={
        rectangle, thick, draw=black, align=center, fill=orange!15,
        minimum height=.3cm, minimum width=2cm, text width=2cm},
    connector/.style={-latex, font=\tiny},
    rectangle connector/.style={
        connector,
        to path={(\tikztostart) -- ++(#1,0pt) \tikztonodes |- (\tikztotarget) },
        pos=0.5
    },
}

\usepackage{color}

\copyrightnotice{\copyright\ IEEE 2021}

\title{Enabling On-Device Training of Speech Recognition Models with Federated Dropout}

\name{\begin{tabular}{c}Dhruv Guliani \qquad Lillian Zhou \qquad Changwan Ryu \qquad Tien-Ju Yang  \\
Harry Zhang  \qquad  Yonghui Xiao \qquad Fran\c{c}oise Beaufays \qquad Giovanni Motta \end{tabular}}
\address{Google LLC, Mountain View, CA, U.S.A. \\
{\{dguliani, lqz, changwan\}@google.com}}
\begin{document}
\bstctlcite{IEEEexample:BSTcontrol}
%
\maketitle
\begin{abstract}
Federated learning can be used to train machine learning models on the edge on local data that never leave devices, providing privacy by default. This presents a challenge pertaining to the communication and computation costs associated with clients' devices. These costs are strongly correlated with the size of the model being trained, and are significant for state-of-the-art automatic speech recognition models.

We propose using federated dropout to reduce the size of client models while training a full-size model server-side. We provide empirical evidence of the effectiveness of federated dropout, and propose a novel approach to vary the dropout rate applied at each layer. Furthermore, we find that federated dropout enables a set of smaller sub-models within the larger model to independently have low word error rates, making it easier to dynamically adjust the size of the model deployed for inference.

\end{abstract}
\section{Introduction}
\label{sec:intro}

End-to-end neural networks have improved to surpass conventional server-side  Automatic Speech Recognition (ASR) systems~\cite{e2e_surpasses_server}. Advances with the Conformer architecture~\cite{gulati2020conformer, e2e_streaming_conformer} have pushed the quality and latency envelopes further, enabling highly performant on-device ASR.

End-to-end neural ASR models can be trained using Federated Learning (FL)~\cite{dguliani_fl_asr}, a privacy-preserving training technique which removes the need to send raw user-data to servers. FL optimization proceeds in synchronous~\emph{rounds} of training~\cite{fl_at_scale}, requiring a set of \emph{clients} (devices) to receive copies of a model at the start of local training and send back model updates for aggregation after optimization.

Unlike other models which have been successfully trained on the edge under FL~\cite{fedemoji, fed_nextword, fed_query, noniid_keyword}, ASR models typically contain over $100$M parameters (see Table~\ref{tab:conformer_sizes}), and it is likely that model sizes will keep increasing~\cite{li_2021_scaling_asr}. This exacerbates communication and computation costs~\cite{dguliani_fl_asr, fl_field_guide} associated with training such models in production. Costs include those of sending and aggregating models, dealing with heterogeneous network dynamics, performing privacy and security computations, and on-device memory.

\begin{table}[htpb]
    \centering
    \small
    \begin{tabular}{c c c}
        \toprule
        \textbf{Conformer} & \textbf{Conf Params} & \textbf{Total Params} \\
        \midrule
        Non-Streaming~\cite{gulati2020conformer} & $107.5$M & $119$M \\
        Streaming~\cite{e2e_streaming_conformer} & $113$M   & $137$M \\
        \bottomrule
    \end{tabular}
    \caption{Architecture of SOTA Conformer models.}
    \label{tab:conformer_sizes}
\end{table}

Federated Dropout~\cite{konecny_fd} (FD) is a technique developed to enable training larger models under FL by reducing the size of models trained on clients. It works by leveraging the key insight from dropout~\cite{srivastava_2014_dropout} that dropping intermediate activations in a network is equivalent to a structural removal of certain rows, columns (and generally, \textit{slices}) of adjacent parameter matrices. An adaptive alternative~\cite{bouacida_2021_adaptive_fd} has also been proposed to better estimate which activations to drop. By reducing the size of client models directly, this technique reduces both communication and computation costs.

In this work, we study the applications of FD to ASR models with the explicit goal of training a full-sized ASR model server-side. We make the following contributions:

\begin{itemize}
    \itemsep-0.1em
    \item We show that FD can be successfully applied to ASR models and provides a useful quality/cost trade-off.
    \item We extend the technique to realistic Google-scale workloads and use varying per-layer dropout rates to achieve better quality with the same size reduction.
    \item We find that training models with FD is an effective way to find well-performing sub-models within a larger model, enabling the size of the model to be reduced for on-device inference. This is useful for deployment on a population of devices having different capabilities.
\end{itemize}

\section{Methodology}
\label{sec:methodology}

\subsection{Model and Data}
We use the largest \textit{Non-Streaming} Conformer described in \cite{gulati2020conformer}, designed for single-domain and short utterances, to conduct initial experiments with FD. This Conformer is trained from scratch under FL with the Librispeech~\cite{libri} corpus split by speaker~\cite{dguliani_fl_asr}. Since we are training these models using FL, we switch from batch normalization to group normalization~\cite{wu_2018_group_norm}, which is more suitable for a federated setting~\cite{noniid_quagmire, adaptive_optimizers}. In doing so, we incur a small penalty in model quality.

We then extend our work to a production-grade Google use case by performing a domain adaptation task using a \textit{Streaming} Conformer. With $137$M trainable parameters, the Streaming Conformer is similar to the one described in~\cite{e2e_streaming_conformer}, but without two-pass re-scoring and cascaded encoders. We train this model to convergence on multi-domain (MD) utterances collected from domains of search, farfield, telephony, YouTube, etc.~\cite{misra21_interspeech, recognizing_long_form} (dataset size in Table~\ref{tab:data_breakdown}). All data are anonymized and
hand-transcribed. In the first pass of training, we withhold one dataset (in this case, the Medium Form (MF) data, which has an average duration of $10.4$ seconds), later refining on data from this domain in a FL simulation. The model is evaluated on a disjoint test set from the MF domain. This is a realistic setting for federated training of ASR models, wherein a well trained server-side model is adapted to a new domain with FL on edge devices. For federated domain adaptation, the data remain anonymously grouped into simulated clients, with each client contributing a uniform amount of data per round. The impact of non-uniform data was previously studied~\cite{dguliani_fl_asr} and is left out of scope for this investigation. Both Conformers studied are encoder-decoder models trained with RNN-T loss.

\begin{table}[h]
    \centering
    \small
    \begin{tabular}{c c c c}
        \toprule
        \textbf{Dataset} & \textbf{Hours} \\
        \midrule
        Multi-domain (MD) & $400$k \\
        \midrule
        Medium-form domain (MF) & $26$k \\
        Medium-form held out (MD-MF) & $374$k \\
        \bottomrule
    \end{tabular}
    \caption{Summary of data sources.}
    \label{tab:data_breakdown}
\end{table}

\begin{table}[]
    \centering
    \begin{tabular}{c | c c}
        \toprule
            \textbf{Portion}
            & \textbf{Non-Streaming} & \textbf{Streaming}  \\
        \midrule
            FeedForward     & $60\%$ & $55\%$ \\
            Attention       & $19\%$ & $15\%$ \\
            Convolution     & $14\%$ & $12\%$ \\
            Decoder         & $3\%$ & $17\%$ \\
            Other           & $4\%$ & $1\%$ \\
        \bottomrule
    \end{tabular}
    \caption{Percentage of parameters that exist in each component of the Conformer models that were studied.}
    \label{tab:conformer_breakdown}
\end{table}

In both experiments, we studied the composition of the models to decide which layers to apply FD to, as the Conformer has various components with different types of layers. We limit our application of FD to just one layer type, and acknowledge that it can be extended to the others in future work. Based on the breakdown in Table~\ref{tab:conformer_breakdown}, we concluded that the FeedForward layers in the encoder were best suited for FD. Dropout rates throughout the paper refer to the ratio of activations dropped within the encoder's FeedForward layers, unless otherwise specified.

\subsection{Federated Dropout}
\label{ssec:feddrop_intro}

Federated Dropout differs from traditional dropout in two main ways. First, the technique is designed to reduce the size of client models for federated training~\cite{konecny_fd} rather than regularizing models. Second, the technique works as a functional transformation during the FL procedure~\cite{fedavg}, wherein smaller client models are \textit{sampled} from the full model for local optimization. Algorithm 1 illustrates this procedure, where each client participating in a round trains a different \textit{sub-model} within the full-size model. In this procedure, the server or client must maintain a reference linking each client's sub-model to the corresponding portions of the full-size model it updates.

Let $K$ be the number of clients participating in a given federated round $r$. Let $w$ be the model parameters, with $W$ being the set of $\{w_1, w_2, \dots, w_k\}$ sub-models that are to be trained by the participating clients in a given round. Let $M$ be the set of $\{m_1, m_2, \dots, m_k\}$ mappings that encode which portion each client's sub-model updates within the full model.

\begin{algorithm}[h]
\caption{\emph{FedDrop}. The $K$ clients are indexed by $k$, rounds are indexed by $r$, and dropout rate is $d$. Shrink($w$, $m$) reduces model weight matrices according to $m$, and Expand($w$, $m$) does the opposite.}
\begin{algorithmic}[1]\label{alg:feddrop}

\For{each round $r$ = 1,2,...}
    \State $M^r$ $\leftarrow$ $GenerateMappings(d)$
    \State $W^{r} \leftarrow Shrink(w^r, M^r)$ \Comment{Set of client models}
    \For{each client $k \in K$ \textbf{in parallel}}
        \State{$\hat{w}^r_k \gets ClientUpdate(k, w^r_k)$}
        \State{$\Delta w^{r}_k = w^r_k - \hat{w}^r_k$}
    \EndFor
    \State $w^{r+1}$ $\leftarrow$ $ServerUpdate(w^r, Expand(\Delta W^{r}, M^r))$
\EndFor
\end{algorithmic}
\end{algorithm}

As can be inferred from Algorithm 1, the selection of mappings $M$ determines how Federated Dropout is applied during each round. We study the two edge cases in which, at each round, either every mapping in $M$ is unique, or all mappings are the same. We call the former Per-Client-Per-Round (PCPR) FD, wherein each participating client updates a different sub-model. We call the latter Per-Round (PR) FD, during which each client updates the same sub-model. This sub-model always changes from one round to the next. Although we hypothesize PCPR would yield the best results, the PR scheme is simpler from a production engineering viewpoint. This is because the PR scheme allows for the identity of each client to be hidden, as the infrastructure need not remember which parts of the full model a particular client updates. This anonymization can be achieved with the PCPR scheme, but requires more engineering effort.

\subsection{Per-Layer Federated Dropout}
\label{ssec:ambient_fd}

We explore making FD more effective by varying the amount of dropout applied across different layers. We target layers for additional dropout using the idea that certain layers of may be \textit{ambient}, or less important to the model's performance~\cite{bengio_2019_ambient_layers}. Ablation experiments involving resetting a model's parameter values to initial/random values can determine which layers are ambient. We apply additional dropout to ambient layers to improve upon the results of uniform FD.

\subsection{Sub-Model Evaluations}
\label{ssec:submodel_evals}

 Other properties of FD are also investigated by sampling and evaluating sub-models from the full size model after training. Sub-models are obtained by removing activations and corresponding neurons in the same way as the FD procedure. The difference is that sub-models are sampled after training from a full-size server-side model, as opposed to dynamically during training (as per Algorithm 1). These models are directly evaluated without any further training.

\section{Experiments}
\label{sec:experiments}

\subsection{Baselines}
\label{ssec:baseline_exps}

Initial experiments established baselines for both the from-scratch and domain adaptation set-ups. These experiments used \textit{SGD} as the client-side optimizer and \textit{Adam}~\cite{adaptive_optimizers} server-side. SpecAugment~\cite{specaug} was applied to utterances in both cases. Additionally, $128$ clients participated in each federated round as informed by previous investigations~\cite{dguliani_fl_asr}. Table~\ref{tab:baselines} summarizes the result for from-scratch training.

\begin{table}[htpb]
    \centering
    \begin{adjustbox}{center, width=\columnwidth-20pt}
    \begin{tabular}{c | c c c c}
    \toprule
        \multirow{2}*{\textbf{Exp.}}
        & \multicolumn{4}{c}{\textbf{WER}} \\
        & \textit{Test} & \textit{TestOther} & \textit{Dev} & \textit{DevOther} \\
    \midrule
        {Non-Streaming Baseline}            & $2.0$ & $4.6$ & $2.3$ & $4.8$ \\
    \bottomrule
    \end{tabular}
    \end{adjustbox}
    \caption{Baseline for from-scratch experiments.\label{tab:baselines}}
\end{table}

For domain adaptation experiments, Table~\ref{tab:streaming_pcpr} shows a \textit{No MF Baseline} which was constructed by training the Streaming model on MD - MF data (Table~\ref{tab:data_breakdown}). This was the starting point for domain adaptation.

\subsection{Training From Scratch}
\label{ssec:initial_exps}

We trained the non-streaming Conformer under FL with varying FD rates, clients per round, and dropout schemes to explore the impact of FD on quality and convergence.

\begin{figure}[htpb]
\centering
\subfigure[Dropout Rate vs. WER]{
    \label{fig:confl_sweep_a}
    \includegraphics[width=0.47\linewidth]{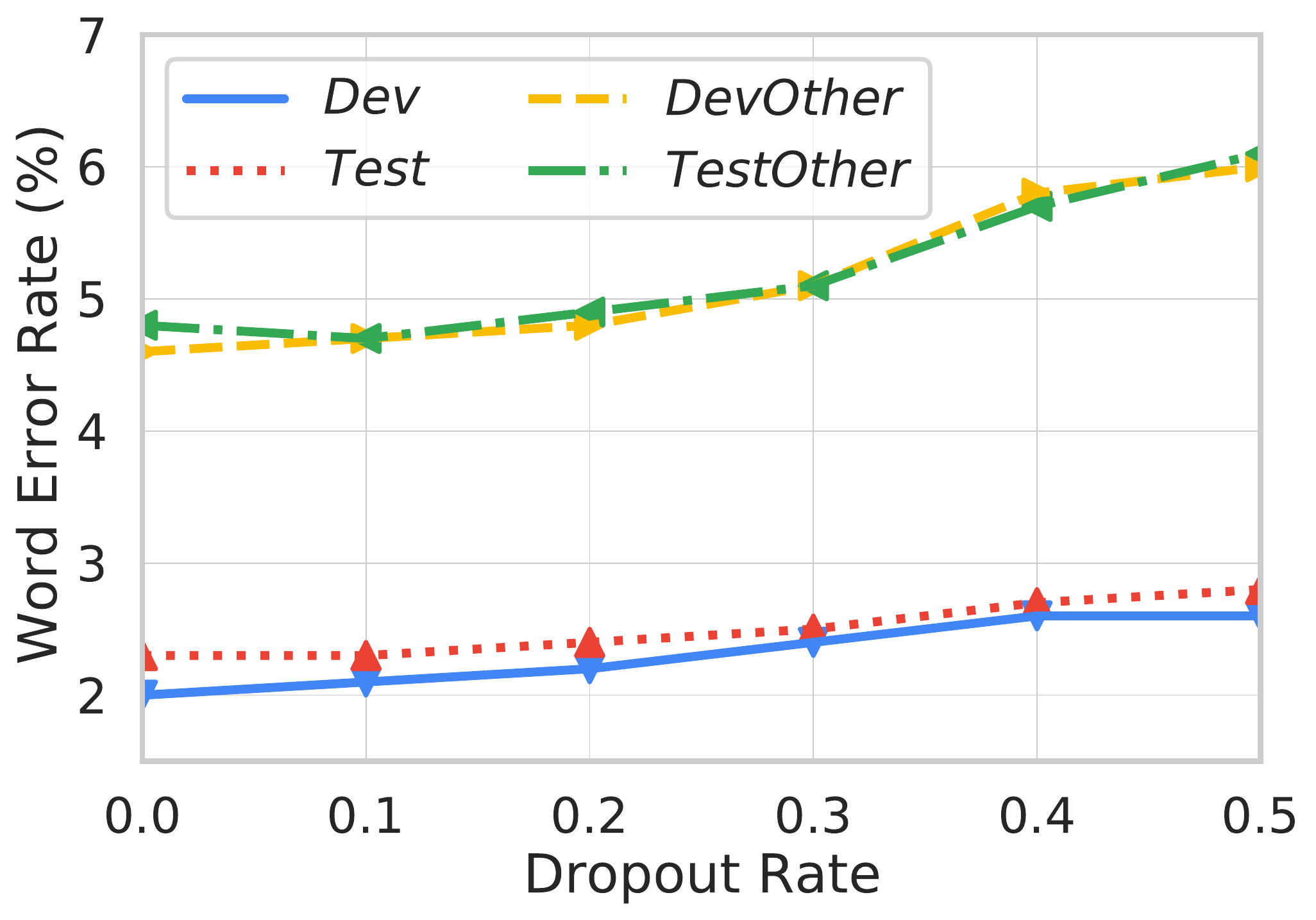}}
\subfigure[Convergence Time]{
    \label{fig:confl_sweep_b}
    \includegraphics[width=0.47\linewidth]{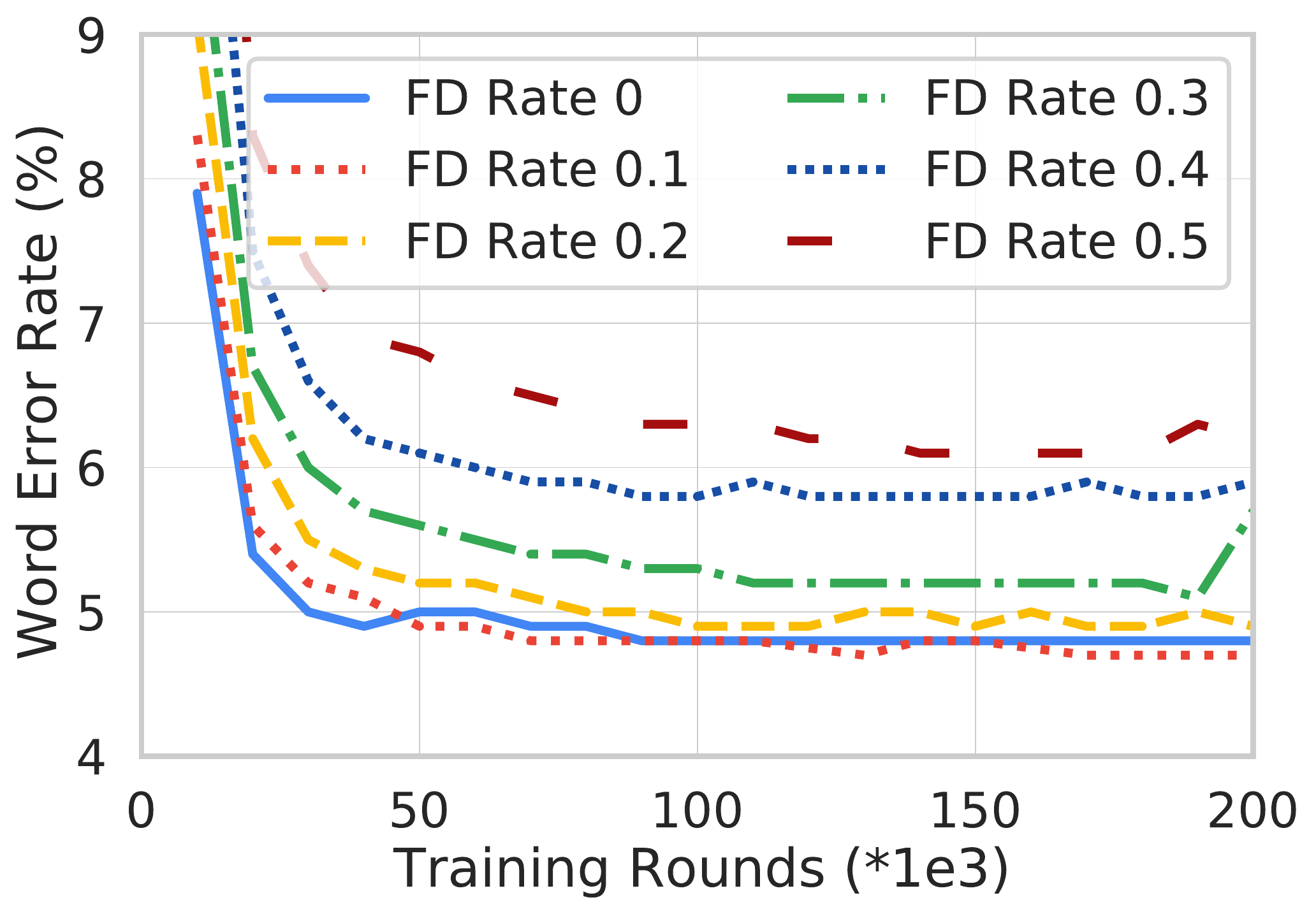}}
\subfigure[PCPR vs. PR]{
    \label{fig:confl_sweep_c}
    \includegraphics[width=0.47\linewidth]{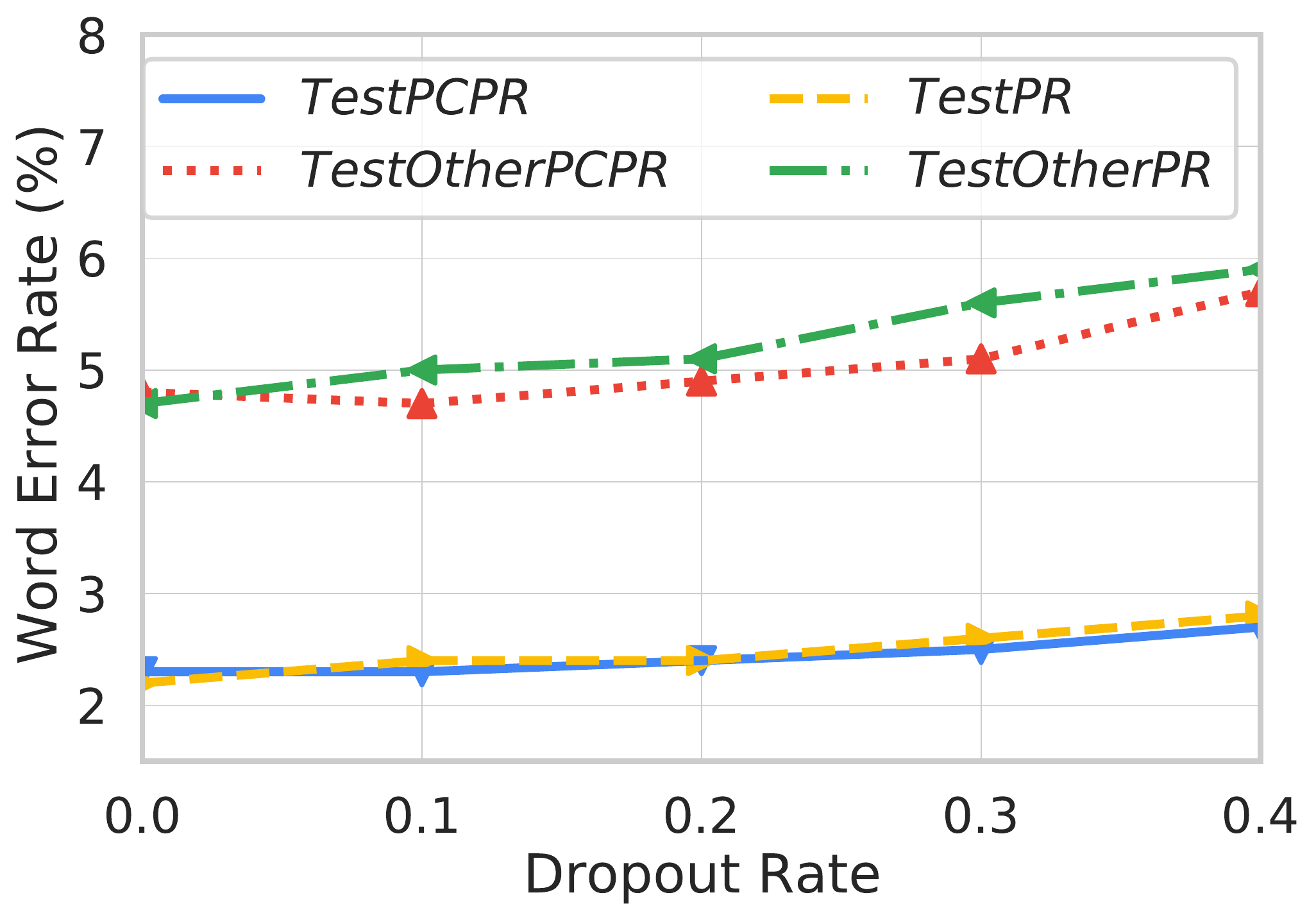}}
\subfigure[Clients Per Round]{
    \label{fig:confl_sweep_d}
    \includegraphics[width=0.47\linewidth]{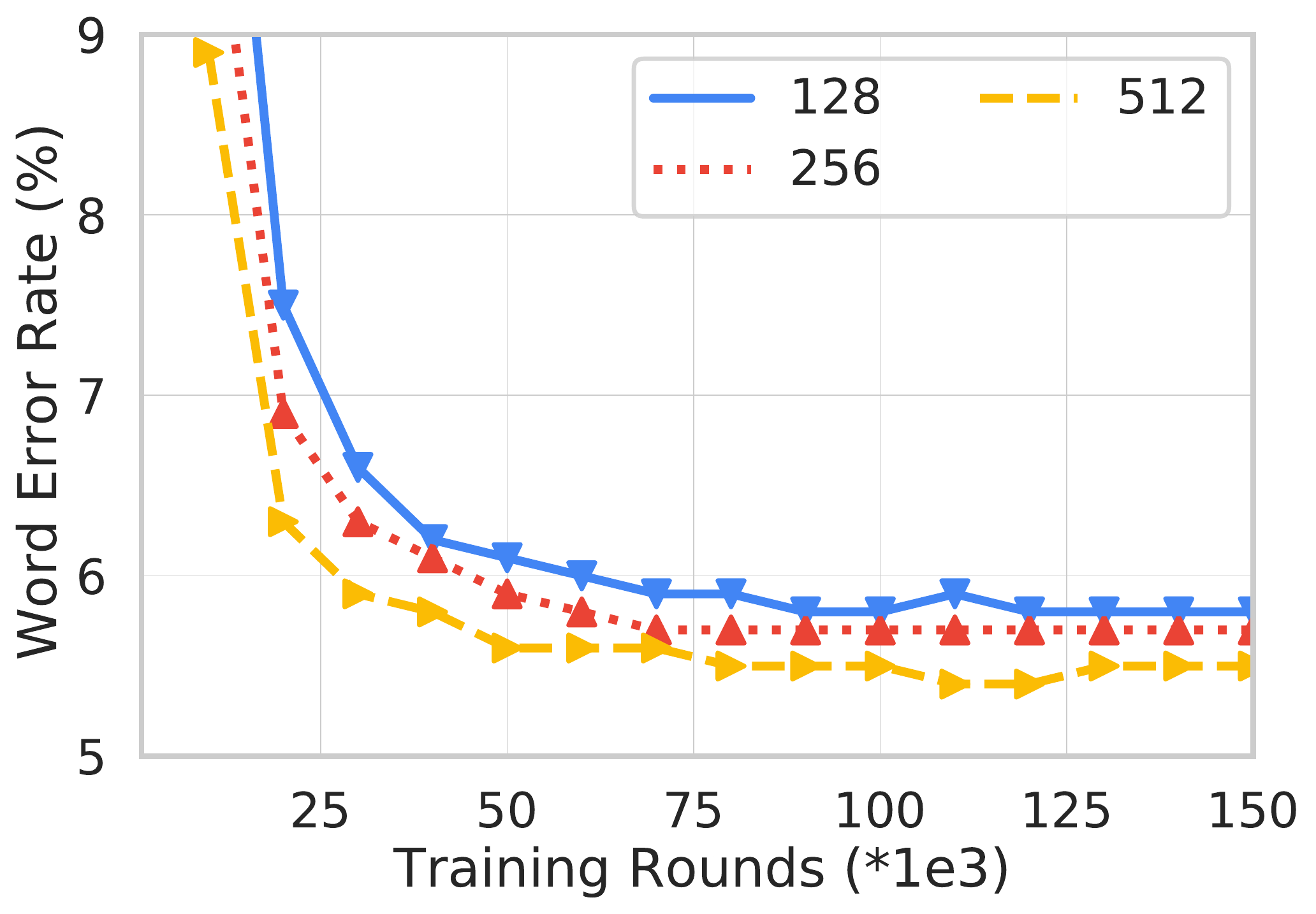}}
\caption{Experimental results showing the impact of Federated Dropout on model quality and convergence with non-IID Librispeech sets. Figures~\ref{fig:confl_sweep_b},~\ref{fig:confl_sweep_c}, and~\ref{fig:confl_sweep_d} report results on the \textit{TestOther} evaluation set, although similar trends were observed across all other evaluation sets. Figure~\ref{fig:confl_sweep_d} reports results using $40\%$ FD.}
\label{fig:confl_experiments}
\end{figure}

Figure~\ref{fig:confl_experiments} shows that increasing dropout rates led to slightly worse models (Figure~\ref{fig:confl_sweep_a}) and slower convergence (Figure~\ref{fig:confl_sweep_b}) across all evaluation sets. Concretely, a $0.5$\% absolute WER regression was observed at the highest dropout rate studied ($50\%$). At this dropout rate, client models were approximately $30$\% smaller in size. Dropout rates up till $20\%$ had minimal quality loss and changes in convergence time, showing a useful trade-off between quality and cost.

The PR scheme (Figure~\ref{fig:confl_sweep_c}) incurred some quality degradation compared to PCPR, but confirmed that the model can be trained effectively by updating only one sub-model in each federated round. With $30\%$ dropout, the PR scheme showed a $0.4\%$ absolute quality degradation in comparison with PCPR. We hence assert this to be a viable and simpler alternative to PCPR with small variations in quality.

Finally, although Figure~\ref{fig:confl_sweep_a} showed that having a higher dropout rate increases convergence time, we empirically show in Figure~\ref{fig:confl_sweep_d} that increasing the number of clients per round can compensate for this. For this particular configuration, model quality was also seen to improve slightly. We therefore claim that quality and convergence time with FD may be improved by tuning hyper-parameters such as clients per round and learning rate.

\subsection{Domain Adaptation}
\label{ssec:fine_tuning_exps}

Domain adaptation experiments (Table~\ref{tab:streaming_pcpr}) showed that PCPR FD could reduce the size of each client model by up to $22$\% while providing a $3$\% relative WER reduction on the previously unseen domain. If a higher quality is desired, client model sizes can be reduced by $11$\% with a $30$\% relative WER improvement on the new domain. Previous results with PR FD also held with the domain adaptation task, wherein PR performed well with some minor quality loss. Higher dropout rates ($50\%$ and greater) resulted in degradation in MF WER from the No MF Baseline.

\begin{table}[t]
    \centering
    \begin{adjustbox}{center, width=\columnwidth-20pt}
    \begin{tabular}{c c | c c c}
    \toprule
        \multirow{2}*{\textbf{Exp.}}
        & \multirow{2}*{\textbf{FD (\%)}}
        & \multicolumn{3}{c}{\textbf{Medium Form WER}} \\
        && \textit{Size Red. (\%)}
        & \textit{PCPR}
        & \textit{PR} \\
    \midrule
        No MF Baseline               &   $0$     & None          & $6.7$           & $6.7$ \\
        MF Domain Ad.       &   $0$     & None          & $4.4$           & $4.4$ \\
    \midrule
        $''$                &   $10$    & $6$           & $4.4$           & $4.4$ \\
        $''$                &   $20$    & $11$          & $4.7$           & $4.7$ \\
        $''$                &   $30$    & $17$          & $5.4$           & $5.5$ \\
        $''$                &   $40$    & $22$          & $6.5$           & $6.6$ \\
    \bottomrule
    \end{tabular}
    \end{adjustbox}
    \caption{Streaming Conformer experiments with FD.\label{tab:streaming_pcpr}}
\end{table}

\subsection{Varying Dropout According to Ambient Properties}
\label{ssec:per_layer}

\begin{figure}[h]
\centering
\includegraphics[width=1\columnwidth]{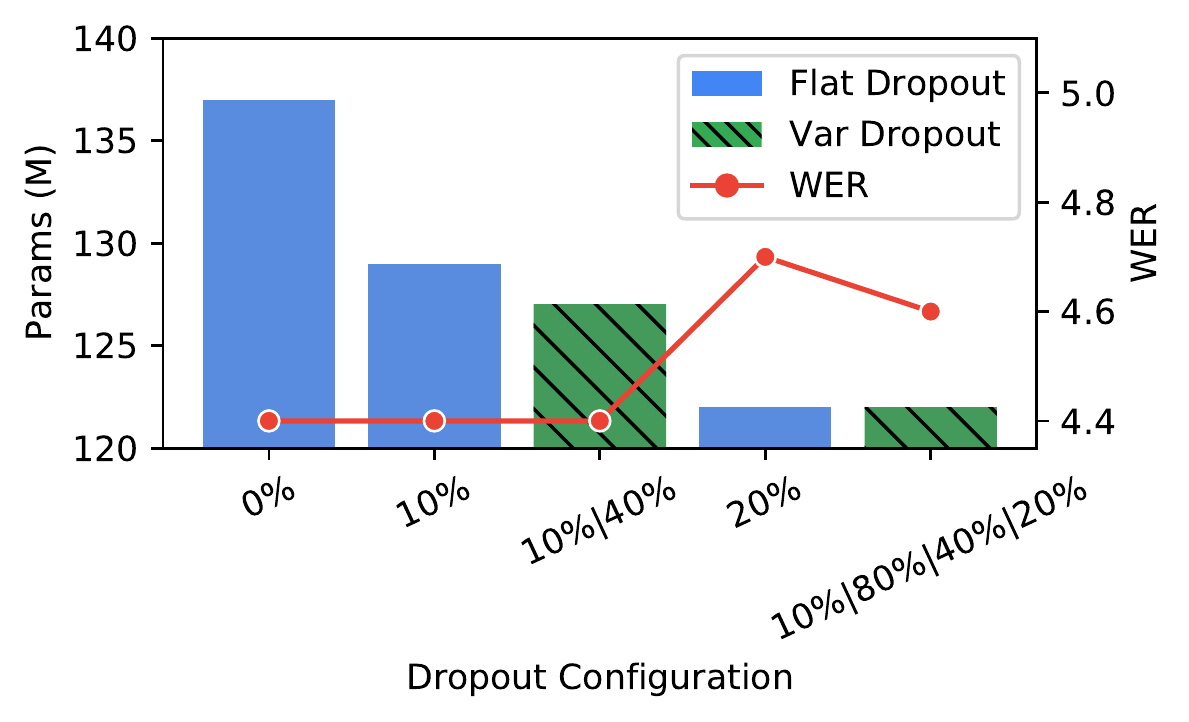}
\caption{Comparison of uniform dropout (solid) and varying dropout (hatched). The labels are of format $d_f|d_1|d_2$... where $d_f$ is flat dropout throughout the model, and $d_n$ denotes dropout applied to the $n^{th}$ most ambient layer.}
\label{fig:per_layer}
\end{figure}

Next, we investigated making FD more effective by using variable dropout rates across the Conformer blocks. We leveraged \textit{ambient layer} findings from \cite{bengio_2019_ambient_layers} to more efficiently target FD. We increased the amount of dropout applied to ambient layers, enabling us to drop more parameters overall without sacrificing model quality. The results in Figure~\ref{fig:per_layer} show that this can provide configurations with smaller client models that give better WER compared to uniform FD throughout the model.

Compared to a flat $10\%$ dropout, we show a setting with variable dropout that gives the same WER, using $1.5\%$ fewer parameters, representing a savings of 2 million parameters. In comparison with a flat $20\%$ dropout, we show a setting with the same number of parameters with a relative $2\%$ WER improvement. We hypothesize that more exhaustive searches of varying dropout rates could yield even more compelling configurations of client model size versus quality.

\subsection{Quality of Sub-Models}
\label{ssec:quality_subnetworks}

Finally, we studied sub-model WERs using the domain adaptation experimental setup. A Streaming Conformer was trained under FL with a $50\%$ FD rate on the MF task, and a second model trained without any FD as the control. Sub-models were sampled ($50$) from each of the models after training, randomly applying $50\%$ dropout for each sample. Table~\ref{tab:submodel} compares the WERs of sampled sub-models. WERs of sub-models from the model trained without FD degraded to catastrophic values higher than $50\%$ on average. In contrast, it was found that FD effectively enabled a flexible set of sub-models within the larger model to achieve a much lower (sub-$10\%$) WER, with lower variance. We conclude that FD can improve the quality of sub-models and enable the same full model to be trimmed down for deployment to various devices with different compute capabilities.

\begin{table}[htpb]
    \centering
    \small
    \begin{tabular}{c | c c}
        \toprule
        \textbf{Exp.} & \textbf{Mean WER} & \textbf{Std. Dev} \\
        \midrule
        {Without FD}    & $50.3$      &   $5.6$ \\
        {With FD}       & $9.5$       &   $0.2$ \\
        \bottomrule
    \end{tabular}
    \caption{WERs of sub-models sampled from the full model.}
    \label{tab:submodel}
\end{table}

\section{Conclusion}
\label{sec:conclusion}

Federated Learning is key to user privacy and ensures that raw user data never leave the device. To leverage this, we must be able to fit model training onto edge devices. End-to-end neural ASR models can contain well over 100 million parameters, creating significant communication and computation cost hurdles on the edge. We argued that Federated Dropout is a promising technique to reduce this cost and explored various configurations to improve its effectiveness. We illustrated a usable quality-cost trade off allowing for client model size reduction between 6-22\%, with WER improvements in a domain adaptation setting ranging from 34-3\% respectively. We also showed that FD causes capable sub-models to form within the full model, allowing the same model to be down-sampled for inference. We hope this work inspires deeper investigations and applications of both client model size reduction and sub-model training.

\section{Acknowledgements}
\label{sec:ack}

We would like to thank Khe Chai Sim and Jakub Kone\v{c}n\'{y}, for providing valuable insights on this paper's contents.

\vfill\pagebreak

\bibliographystyle{IEEEtran}
\bibliography{refs}

\end{document}